\documentclass[11pt,a4paper]{article}
\usepackage[hyperref]{emnlp2020}
\usepackage{times}
\usepackage{latexsym}

\usepackage{multirow}
\usepackage{amsfonts}
\usepackage{subfig}
\usepackage{graphicx} 
\usepackage{array}
\usepackage{float}
\usepackage{url}
\usepackage{amsmath}
\usepackage{enumitem}
\usepackage{amssymb}
\usepackage{cleveref}
\crefformat{section}{\S#2#1#3}
\crefformat{subsection}{\S#2#1#3}
 
\usepackage{booktabs}
\usepackage{mathtools}

\newcommand\blfootnote[1]{%
\begingroup 
\renewcommand\thefootnote{*}\footnotetext{#1}%
\addtocounter{footnote}{0}%
\endgroup 
}

\usepackage{microtype}

\aclfinalcopy 

\title{VD-BERT: A Unified Vision and Dialog Transformer with BERT}

\author{Yue Wang$^{1*}$,~Shafiq Joty$^{2}$,~Michael R. Lyu$^{1}$,~Irwin King$^{1}$,~Caiming Xiong$^{2}$,~and Steven C.H. Hoi$^{2}$\\
$^{1}$ Department of Computer Science and Engineering\\
The Chinese University of Hong Kong, HKSAR, China \\
$^{2}$Salesforce Research\\
$^{1}$\texttt{\{yuewang,lyu,king\}@cse.cuhk.edu.hk}\\
$^{2}$\texttt{\{sjoty,cxiong,shoi\}@salesforce.com}\\}

\date{}

\begin{document}
\maketitle
\blfootnote{This work was mainly
done when Yue Wang was an intern at Salesforce Research Asia, Singapore.}
\begin{abstract}
Visual dialog is a challenging vision-language task, where a dialog agent needs to answer a series of questions through reasoning on the image content and dialog history. Prior work has mostly focused on various attention mechanisms to model such intricate interactions. By contrast, 
in this work, we propose {VD-BERT}, a simple yet effective framework of unified vision-dialog Transformer that leverages the pretrained BERT language models for {V}isual {D}ialog tasks.
The model is \emph{unified} in that (1) it captures all the  interactions between the image and the multi-turn dialog using a single-stream Transformer encoder, and (2) it supports both answer ranking and answer generation seamlessly through the same architecture.
More crucially, we  adapt  BERT for the effective fusion of vision and dialog contents via \textit{visually grounded} training.
Without the need of pretraining on external vision-language data, our model yields new state of the art, achieving the top position in both single-model and ensemble settings ($74.54$ and $75.35$ NDCG scores) on the visual dialog leaderboard. Our code and pretrained models are released at \url{https://github.com/salesforce/VD-BERT}.
\end{abstract}

\section{Introduction}
Visual Dialog (or VisDial) aims to build an AI agent that can answer a human's questions about visual content in a natural conversational setting \cite{DBLP:conf/cvpr/DasKGSYMPB17}.
Unlike the traditional single-turn Visual Question Answering (VQA) \cite{DBLP:conf/iccv/AntolALMBZP15}, the agent in VisDial requires to answer questions through multiple rounds of interactions together with  visual content understanding. 

\begin{figure}
\centering
\includegraphics[scale=0.45]{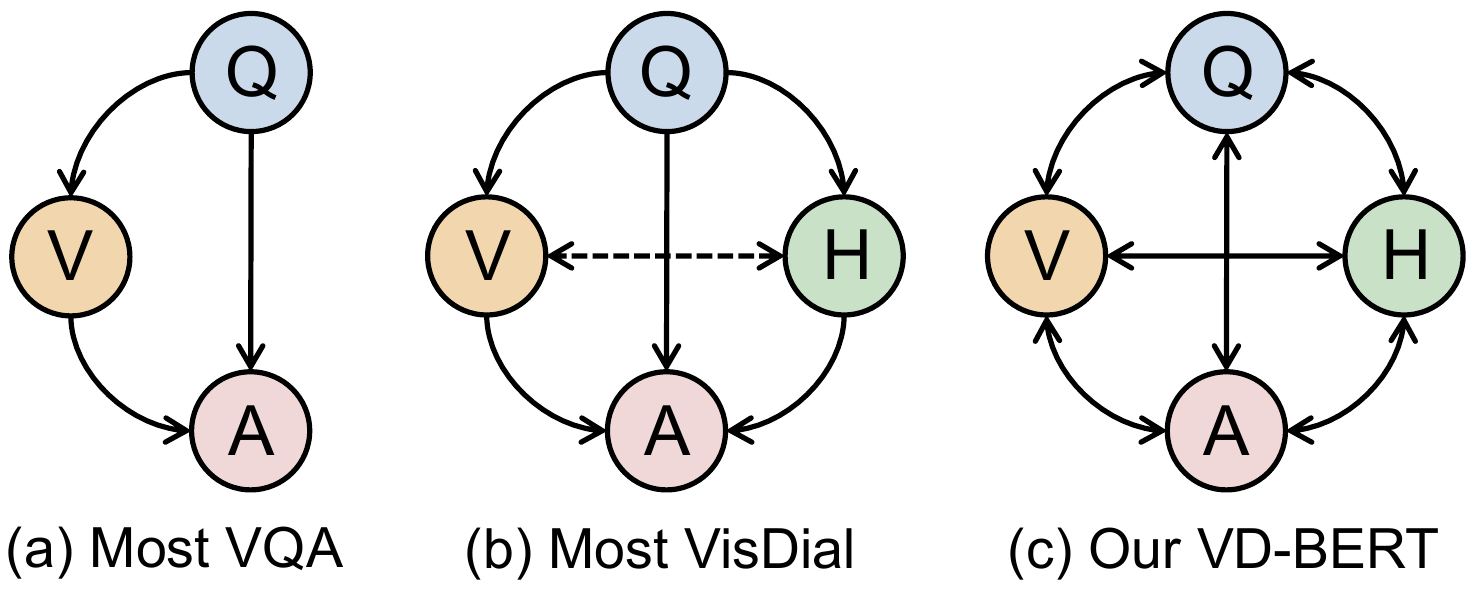}
\vspace{-0.3em}
\caption{Attention flow direction illustration. V: vision, H: dialog history, Q: question, A: answer. The arrow denotes the attention flow direction and the dashed line represents an optional connection.}\label{fig:attn_direction}
\vspace{-0.5em}
\end{figure}

The primary research direction in VisDial has been mostly focusing on developing various attention mechanisms~\cite{DBLP:journals/corr/BahdanauCB14} for a better fusion of vision and dialog contents.
Compared to VQA that predicts an answer based only on the question about the image  (Figure~\ref{fig:attn_direction}(a)), VisDial needs to additionally consider the dialog history.
Typically, most of previous work~\cite{DBLP:conf/cvpr/NiuZZZLW19,DBLP:conf/acl/GanCKLLG19,DBLP:conf/emnlp/KangLZ19} uses the question as a query to attend to relevant image regions and dialog history,
where their interactions are usually exploited to obtain better visual-historical cues for predicting the answer. In other words, the attention flow in these methods is \emph{unidirectional} -- from question to the other entities (Figure~\ref{fig:attn_direction}(b)).

By contrast, in this work, we allow for  \emph{bidirectional} attention flow between all the entities using a unified Transformer~\cite{DBLP:conf/nips/VaswaniSPUJGKP17} encoder, as shown in Figure~\ref{fig:attn_direction}(c). In this way, all the entities simultaneously play the role of an ``information seeker'' (query) and an ``information provider'' (key-value), thereby fully unleashing the potential of attention similar to~\citet{DBLP:conf/cvpr/SchwartzYHS19}. We employ the Transformer as the encoding backbone due to its powerful representation learning capability exhibited in pretrained language models like BERT~\cite{DBLP:conf/naacl/DevlinCLT19}.
Inspired by its recent success in  vision-language pretraining, we further extend BERT to achieve simple yet effective fusion of vision and dialog contents in VisDial tasks.

Recently some emerging work has attempted to adapt BERT for multimodal tasks~\cite{DBLP:conf/iccv/SunMV0S19,DBLP:conf/nips/LuBPL19,tan2019lxmert,DBLP:conf/aaai/ZhouPZHCG20}.
They often use self-supervised objectives to pretrain BERT-like models on large-scale external vision-language data and then fine-tune on downstream tasks. This has led to compelling results in tasks such as VQA, image captioning, image retrieval~\cite{DBLP:journals/tacl/YoungLHH14}, and visual reasoning~\cite{DBLP:conf/acl/SuhrZZZBA19}. However, it is still unclear how visual dialog may benefit from such vision-language pretraining due to its unique multi-turn conversational structure.
Specifically, each image in the VisDial dataset is associated with up to $10$ dialog turns, which contain much longer contexts than either VQA or image captioning.

In this paper, we present VD-BERT, a novel unified vision-dialog Transformer framework for VisDial tasks. Specifically, we first encode the image into a series of detected objects and feed them into a  Transformer encoder together with the image caption and multi-turn dialog. 
We initialize the encoder with  BERT for better leveraging the pretrained language representations.
To effectively fuse features from the two modalities, we make use of two  \textit{visually grounded} training objectives -- Masked Language Modeling (MLM) and Next Sentence Prediction (NSP). Different from the original MLM and NSP in BERT, we additionally take the visual information into account when predicting the masked tokens or the next answer. 

VisDial models have been trained in one of two settings: discriminative or generative. In the discriminative setting, the model ranks a pool of  answer candidates, whereas the generative setting additionally allows the model to generate the answers. Instead of employing two types of decoders like prior work, we rely on a unified Transformer architecture with two different self-attention masks~\cite{DBLP:conf/nips/00040WWLWGZH19} to seamlessly  support both settings.
During inference, our VD-BERT either ranks the answer candidates according to their NSP scores or generates the answer sequence by recursively applying the MLM operations. 
We further fine-tune our model on dense annotations that specify the relevance score for each answer candidate with a ranking optimization module. 

\noindent In summary, we make the following contributions: 
\begin{itemize}[leftmargin=*]
\vspace{-0.5em}
\itemsep0em
    \item To the best of our knowledge, our work serves as one of the first attempts to explore pretrained language models for visual dialog. {We showcase that BERT can be effectively adapted to this task with simple visually grounded training for capturing the intricate vision-dialog interactions.} Besides, our VD-BERT is the first unified model that  supports both discriminative and generative training settings  without  explicit decoders.
    \item Without  pretraining on external vision-language data, our model yields new state-of-the-art results in  discriminative setting and promising results in generative setting on  VisDial  benchmarks (\cref{ssec:main_res}). 
    \item We conduct extensive experiments not only to analyze how our model performs with various training aspects (\cref{ssec:analysis}) and fine-tuning on dense annotations (\cref{ssec:dense}), but also to interpret it via  attention visualization  (\cref{ssec:attn_vis}),  shedding light on future transfer learning research for VisDial tasks.
   
\end{itemize}

\section{Related Work}
\paragraph{Visual Dialog.} 
The Visual Dialog task has been recently proposed by \citet{DBLP:conf/cvpr/DasKGSYMPB17}, where a dialog agent needs to answer a series of questions grounded by an image. 
It is one of the most challenging vision-language tasks that require not only to understand the image content according to texts, but also to reason through the dialog history. Previous work~\cite{DBLP:conf/nips/LuKYPB17,DBLP:conf/nips/SeoLHS17,DBLP:conf/cvpr/Wu0S0H18,DBLP:conf/eccv/KotturMPBR18,DBLP:conf/aaai/JiangYQZZ0W20,DBLP:conf/iccv/YangZZ19,DBLP:conf/cvpr/GuoXT19,DBLP:conf/cvpr/NiuZZZLW19} focuses on developing a variety of attention mechanisms to model the interactions among entities including image, question, and dialog history. For example, \citet{DBLP:conf/emnlp/KangLZ19} proposed DAN, a dual attention module to first refer to relevant contexts in the dialog history, and then find indicative image regions. ReDAN, proposed by \citet{DBLP:conf/acl/GanCKLLG19}, further explores the interactions between image and dialog history via multi-step reasoning.

\begin{figure*}[t]
\centering
\includegraphics[scale=0.45]{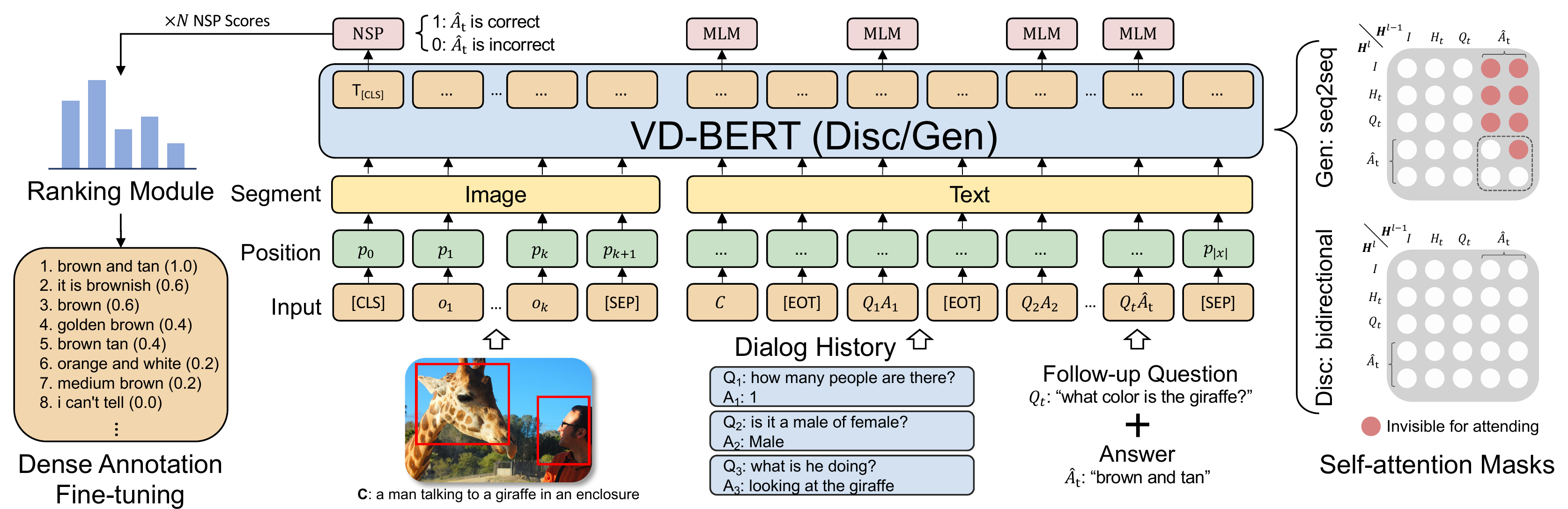}
\vspace{-0.3em}
\caption{The model architecture of our unified VD-BERT for both discriminative and generative settings.}\label{fig:framework}
\vspace{-0.7em}
\end{figure*}


Different from them, we rely on the self-attention mechanism within a single-stream Transformer encoder to capture such complex interactions in a unified manner and derive a ``holistic'' contextualized representation for all the  entities.
Similar to this, \citet{DBLP:conf/cvpr/SchwartzYHS19} proposed FGA, a general factor graph attention that can model interactions between any two entities but in a pairwise manner.
There is recent work~\cite{DBLP:journals/corr/abs-1911-11390, DBLP:conf/acl/AgarwalBLKR20} also applying the Transformer to model the interactions among many entities. However, their models neglect the important early interaction of the answer entity and cannot naturally leverage the pretrained language representations from BERT like ours.

Regarding the architecture, our model mainly differs from previous work in two facets:
first, unlike most prior work that considers answer candidates only at the final similarity computation layer, our VD-BERT integrates each answer candidate at the input layer to  enable its early and deep fusion with other entities, similar to~\citet{DBLP:conf/cvpr/SchwartzYHS19};
second,  existing models adopt an encoder-decoder framework~\cite{DBLP:conf/nips/SutskeverVL14} with two types of decoder for the discriminative and generative settings separately, while
 we instead adopt a unified Transformer encoder with two different self-attention masks~\cite{DBLP:conf/nips/00040WWLWGZH19} to seamlessly support both settings without extra decoders.

\paragraph{Pretraining in Vision and Language.}
Pretrained language models like
BERT~\cite{DBLP:conf/naacl/DevlinCLT19}  have boosted  performance greatly in a broad set of NLP tasks.
In order to benefit from the pretraining, 
there are many recent attempts to extend BERT for vision and language pretraining.
They typically employ the Transformer encoder as the backbone with either a two-stream architecture to encode text and image independently such as ViLBERT~\cite{DBLP:conf/nips/LuBPL19} and LXMERT~\cite{tan2019lxmert}, or a single-stream
architecture to encode both text and image together, such as  B2T2~\cite{DBLP:conf/emnlp/AlbertiLCR19},
Unicoder-VL~\cite{DBLP:conf/aaai/LiDFGJ20},
VisualBERT~\cite{DBLP:journals/corr/abs-1908-03557},  VL-BERT~\cite{DBLP:conf/iclr/SuZCLLWD20}, and UNITER~\cite{chen2019uniter}. 
Our VD-BERT belongs to the second group.
These models yield prominent improvements mainly on  vision-language understanding tasks like VQA, image retrieval~\cite{DBLP:journals/tacl/YoungLHH14}, and visual reasoning~\cite{DBLP:conf/acl/SuhrZZZBA19,DBLP:conf/cvpr/ZellersBFC19}.

More recently, \citet{DBLP:conf/aaai/ZhouPZHCG20} proposed VLP which also allows generation using a unified Transformer with various self-attention masks~\cite{DBLP:conf/nips/00040WWLWGZH19}.  
Their model was proposed for VQA and image captioning.
Our model is inspired by VLP and specifically tailored for the visual dialog task.
Most closely related to this paper is the concurrent work VisDial-BERT by~\citet{DBLP:journals/corr/abs-1912-02379}, who also employ  pretrained models (i.e., ViLBERT) for visual dialog.
Our work has two major advantages over VisDial-BERT: first, VD-BERT supports both discriminative and generative settings while theirs is restricted to only the discriminative setting; second, we do not require to pretrain on large-scale external vision-language datasets like theirs and still yield better performance (\cref{ssec:main_res}).

\section{The VD-BERT Model}

We first formally describe the visual dialog task. Given a question $Q_t$ grounded on an image $I$ at $t$-th turn, as well as its dialog history formulated as $H_t=\{C, (Q_1,A_1),...,(Q_{t-1},A_{t-1})\}$ (where $C$ denotes the image caption), the agent is asked to predict its answer $A_t$ by ranking a list of $100$ answer candidates $\{\hat{A}_t^1, \hat{A}_t^2,...,\hat{A}_t^{100}\}$.
In general, there are two types of decoder to predict the answer: a \textit{discriminative} decoder that \textit{ranks} the answer candidates and is trained with a cross entropy loss, or a \textit{generative} decoder that \textit{synthesizes} an answer and is trained with a maximum log-likelihood loss.

Figure~\ref{fig:framework} shows the overview of our approach. First, we employ a unified  vision-dialog  Transformer to encode both the image and dialog history, where we  append an answer candidate $\hat{A}_t$ in the input to model their interactions in an early fusion manner (\cref{ssec:enc_pt}).
Next, we adopt  visually grounded MLM and NSP objectives to train the model for effective vision and dialog fusion  using two types of self-attention masks -- bidirectional and seq2seq. This allows our unified model to work in both discriminative and  generative settings (\cref{ssec:learning}). Lastly, we devise a ranking optimization module to further fine-tune on the dense annotations (\cref{ssec:rank}).

\subsection{Vision-Dialog Transformer Encoder}\label{ssec:enc_pt}

\paragraph{Vision Features.} Following previous work, we employ Faster R-CNN~\cite{DBLP:conf/nips/RenHGS15} pretrained on Visual Genome~\cite{DBLP:journals/ijcv/KrishnaZGJHKCKL17} to extract the object-level vision features. Let $O_I = \{o_1,...,o_k\}$ denote the vision features for an image $I$, where each object feature $o_i$ is a $2048$-d Region-of-Interest (RoI) feature and $k$ is the number of the detected objects (fixed to $36$ in our setting). As there is no natural orders among these objects, we adopt normalized bounding box coordinates as the spatial location. Specifically, let $(x_1, y_1)$ and $(x_2, y_2)$ be the coordinates of the bottom-left and top-right corner of the $i$-th object, its  location information is encoded into a $5$-d vector: $p_i = (\frac{x_1}{W}, \frac{y_1}{H},\frac{x_2}{W}, \frac{y_2}{H}, \frac{(x_2-x_1)(y_2-y_1)}{WH})$, where $W$ and $H$ respectively denote the width and height of the input image, and the last element is the relative area of the object. 
We  extend $p_i$ with its class id and confidence score for a richer representation.

\vspace{-0.2em}
\paragraph{Language Features.} 
We pack all the textual elements (caption and multi-turn dialog) into a long sequence. We employ WordPiece tokenizer~\cite{wu2016google} to split it into a word sequence $\mathbf{w}$, where each word is embedded with an absolute positional code following~\citet{DBLP:conf/naacl/DevlinCLT19}.

\vspace{-0.2em}
\paragraph{Cross-Modality Encoding.} To feed both image and text into the Transformer encoder, we integrate the image objects with language elements into a whole input sequence.
Similar to BERT, we use special tokens like \texttt{[CLS]} to denote the beginning of the sequence, and \texttt{[SEP]} to separate the two  modalities. 
Moreover, to inject the multi-turn dialog structure into the model, we utilize a special token  \texttt{[EOT]} to denote \textit{end of turn}~\cite{DBLP:journals/corr/abs-1908-04812}, which informs the model when the dialog turn ends.
As such, we prepare the input sequence into the format
as $\mathbf{x}$ $=$ (\texttt{[CLS]}, $o_1,...,o_k$, \texttt{[SEP]}, $C$, \texttt{[EOT]}, $Q_1A_1$, \texttt{[EOT]}, ...,  $Q_t\hat{A}_t$, \texttt{[SEP]}). 
To notify the model for the answer prediction, we further insert a \texttt{[PRED]}  token between the $Q_t\hat{A}_t$ pair.
Finally, each input token embedding is combined with its position embedding and segment embedding ($0$ or $1$, indicating whether it is image or text) with layer normalization~\cite{DBLP:journals/corr/BaKH16}.

\vspace{-0.2em}
\paragraph{Transformer Backbone.}
We denote the embedded vision-language inputs as $\mathbf{H}^0=[\mathbf{e}_1, ..., \mathbf{e}_{|\mathbf{x}|}]$ and then encode them into multiple levels of contextual representations $\mathbf{H}^l=[\mathbf{h}_1^l, ...,\mathbf{h}^l_{|\mathbf{x}|}]$ using $L$-stacked  Transformer blocks, where the $l$-th Transformer block is denoted as $\mathbf{H}^l=\text{Transformer}(\mathbf{H}^{l-1}), l\in[1, L]$. Inside each Transformer block, the previous layer's output $\mathbf{H}^{l-1}\in\mathbb{R}^{|\mathbf{x}|\times d_h}$ is aggregated using the multi-head self-attention~\cite{DBLP:conf/nips/VaswaniSPUJGKP17}:
\begin{equation}
    \mathbf{Q}=\mathbf{H}^{l-1}\mathbf{W}_l^{Q}, \mathbf{K}=\mathbf{H}^{l-1}\mathbf{W}_l^{K}, \mathbf{V}=\mathbf{H}^{l-1}\mathbf{W}_l^{V},
    \vspace{-0.3in}
\end{equation}
\begin{equation}\label{eq:mask}
    \mathbf{M}_{ij}= \begin{cases}
      0, & \text{allow to attend},\\
      -\infty, & \text{prevent from attending},\\
    \end{cases}  
     \vspace{-0.1in}
\end{equation}
\begin{equation}
    \mathbf{A}_l=\text{softmax}(\frac{\mathbf{Q}\mathbf{K}^T}{\sqrt{d_k}}+\mathbf{M})\mathbf{V},
\end{equation}
where $\mathbf{W}_l^{Q}, \mathbf{W}_l^{K}, \mathbf{W}_l^{V}\in\mathbb{R}^{d_h\times d_k}$ are learnable weights for computing the queries, keys, and values respectively, and $\mathbf{M}\in\mathbb{R}^{|\mathbf{x}|\times|\mathbf{x}|}$ is the self-attention mask that determines whether tokens from two  layers can attend each other.
Then $\mathbf{A}_l$ is passed into a feedforward layer to compute $\mathbf{H}^l$ for the next layer.

\subsection{Visually Grounded Training Objectives} \label{ssec:learning}
We use two \textit{visually grounded} training objectives---masked language modeling (MLM) and next sentence prediction (NSP) to train our VD-BERT. 
Particularly,
we aim to capture dense interactions among both inter-modality  (i.e., image-dialog) and intra-modality (i.e., image-image, dialog-dialog).

Similar to MLM in BERT, $15\%$ tokens in the text segment
(including special tokens like \texttt{[EOT]} and \texttt{[SEP]}) 
are randomly masked out and replaced with a special token \texttt{[MASK]}.
The model is then required  to recover them based   not only on the surrounding tokens $\mathbf{w}_{\backslash m}$ but also on the image $I$:
\begin{equation}\label{eq:MLM}
  \mathcal{L}_{MLM} = -E_{(I, \mathbf{w})\sim D}\log P(w_m| \mathbf{w}_{\backslash m}, I),
\end{equation}
where $w_m$ refers to the masked token and $D$ denotes the training set.
Following~\citet{DBLP:conf/aaai/ZhouPZHCG20}, we do not conduct similar masked object/region modeling in the image segment. 

As for NSP, instead of modeling the relationship between two sentences (as in BERT) or the matching of an image-text pair (as in other vision-language pretraining models like ViLBERT), VD-BERT aims to predict whether the appended answer candidate $\hat{A}_t$ is correct or not based on the joint understanding of the image and dialog history:
\begin{equation}\label{eq:NSP}
  \mathcal{L}_{NSP} = -E_{(I, \mathbf{w})\sim D}  \log P(y|S(I, \mathbf{w})),
\end{equation}
where $y\in\{0,1\}$ indicates whether $\hat{A}_t$ is correct, and  $S(\cdot)$ is a binary classifier to predict the probability based on the \texttt{[CLS]} representation {T\textsubscript{\texttt{[CLS]}}} at the final layer. 
Below we introduce the discriminative and generative settings of VD-BERT.

\vspace{-0.2em}
\paragraph{Discriminative Setting.} 
For training in the discriminative setting, we transform the task of selecting an answer into a pointwise binary classification problem. 
Specifically, we sample an answer $\hat{A}_t$ from the candidate pool and append it to the input sequence, and ask the NSP head to distinguish whether the sampled answer is correct or not.
We employ the \textit{bidirectional} self-attention mask to allow all the tokens to attend to each other by setting the mask matrix $\mathbf{M}$ in Eq.~(\ref{eq:mask}) to all $0$s.
To avoid imbalanced class distribution,  we keep the ratio of positive and negative instances to 1:1 in each epoch.
To encourage the model to penalize more on negative instances, we randomly resample a negative example from the pool of $99$ negatives w.r.t. every positive one at different epochs.
During inference, we rank the answer candidates according to the positive class score  of their NSP heads.

\vspace{-0.2em}
\paragraph{Generative Setting.} In order to  autoregressively generate an answer, we also train VD-BERT with the \textit{sequence-to-sequence} (seq2seq) self-attention mask~\cite{DBLP:conf/nips/00040WWLWGZH19}. For this, we divide the input sequence to each Transformer block into two subsequences, \emph{context} and \emph{answer}:
\begin{equation}\label{eq:divide}
  \mathbf{x}\triangleq(I, \mathbf{w})=(\underbrace{I, H_t, Q_t,}_{\text{context}} \hat{A}_t).
\end{equation}
We allow tokens in the context to be fully visible for attending by setting  the left part of $\mathbf{M}$ to all $0$s.
For the answer sequence, we mask out (by setting $-\infty$ in $\mathbf{M}$) the ``future" tokens to get  autoregressive attentions (see the red dots in Figure~\ref{fig:framework}).

During inference, we rely on the same unified Transformer encoder with sequential MLM operations without an explicit decoder. 
Specifically, we recursively append a \texttt{[MASK]} token to the end of the sequence to trigger a one-step prediction and then replace it with the predicted token for the next token prediction.
The decoding process is based on greedy sampling and terminated when a \texttt{[SEP]} is emitted, and the resulting log-likelihood scores will be used for ranking the answer candidates.

\subsection{Fine-tuning with Rank Optimization} \label{ssec:rank}
As some answer candidates may be semantically similar (e.g.,  ``brown and tan'' vs ``brown'' in Figure \ref{fig:framework}), VisDial v$1.0$ additionally provides dense annotations that specify real-valued relevance scores for the $100$ answer candidates, $[s_1,...,s_{100}]$ with $s_i\in [0,1]$. To fine-tune on this, we combine the NSP scores from the model for all answer candidates together into a vector $[p_1,...,p_{100}]$. 

As dense annotation fine-tuning is typically a Learning to Rank (LTR) problem, we can make use of some ranking optimization methods. 
After
comparing various methods in Table~\ref{tabs:ablation}(c),
we adopt ListNet~\cite{DBLP:conf/icml/CaoQLTL07} with the top-$1$ approximation as the ranking module for  VD-BERT:
\begin{equation}
    \mathcal{L}_{ListNet}=-\sum_{i=1}^{N} f(s_i)\log(f(p_i)),
\end{equation}
\vspace{-0.2in}
\begin{equation}
  f(x_i)=\frac{\exp{(x_i)}}{\sum_{j=1}^N\exp{(x_j)}},~~i=1,...,N.
\end{equation}
For training efficiency, we sub-sample the candidate list and use only  $N=30$ answers  (out of 100) for each instance. To better leverage the contrastive signals from the dense annotations, the sub-sampling method first picks randomly the candidates with non-zero relevance scores, and then it picks the ones from zero scores (about $12\%$ of candidates are non-zero on average).

\section{Experimental Setup}\label{sec:exp_setup}
\paragraph{Datasets.}
We evaluate our model on the VisDial v$0.9$ and v$1.0$ datasets~\cite{DBLP:conf/cvpr/DasKGSYMPB17}.
Specifically, v$0.9$ contains a training set of 82,783 images and a validation set of 40,504 images. The v$1.0$ dataset combines the training and validation sets of v$0.9$ into one training set and adds another 2,064 images for validation and 8,000 images for testing (hosted blindly in the task organizers' server).  Each image is associated with one caption and 10 question-answer pairs.
For each question, it is paired with a list of 100 answer candidates, one of which is regarded as the correct answer.

For the v$1.0$ validation split and a part of v$1.0$ train split (2,000 images), extra dense annotations for the answer candidates are provided to make the evaluation more reasonable. 
The dense annotation specifies a relevance score for each answer candidate based on the fact that some candidates with similar semantics to the ground truth answer can also be considered as correct or partially correct, e.g., ``brown and tan'' and ``brown'' in Figure~\ref{fig:framework}.

\vspace{-0.2em}
\paragraph{Evaluation Metric.}
Following~\citet{DBLP:conf/cvpr/DasKGSYMPB17}, we evaluate our model using the ranking metrics like Recall@K (K  $\in \{1,5,10\}$), Mean Reciprocal Rank (MRR), and Mean Rank, where only one  answer is considered as correct.
Since the 2018 VisDial challenge (after the acquisition of  dense annotations), NDCG metric that considers the relevance degree of each answer candidate, has been adopted as the main metric to determine the winner.

\vspace{-0.2em}
\paragraph{Configurations.}
We use BERT\textsubscript{BASE} as the backbone, which consists of $12$ Transformer blocks, each with 12 attention heads and a hidden state dimensions of $768$.   
We keep the max input sequence length (including $36$ visual objects) to $250$.
We use Adam~\cite{DBLP:journals/corr/KingmaB14} with an initial learning rate of $3e-5$ and a batch size of $32$ to train our model.
A linear learning rate decay schedule with a warmup of $0.1$ is employed.
We first train VD-BERT for $30$ epochs on a cluster of $4$ V$100$ GPUs with $16$G memory using MLM and NSP losses (with equal coefficients).
Here we only utilize one previous dialog turn for training efficiency.
For instances where the appended answer candidate is incorrect, we do not conduct MLM on the answer sequence to reduce the noise introduced by the negative samples.
After that, we  train for another $10$ epochs with full dialog history using either NSP in the discriminative setting or MLM on the answer sequence in the generative setting.
For dense annotation fine-tuning in the discriminative setting, we train with the ListNet loss for $5$ epochs.

\section{Results and Analysis}\label{sec:exp}

We first compare VD-BERT with state-of-the-art models on VisDial datasets (\cref{ssec:main_res}). Then we conduct ablation studies to examine various aspects of our model (\cref{ssec:analysis}), followed by an in-depth analysis of fine-tuning on dense annotations~(\cref{ssec:dense}).
Lastly, we interpret how it attains the effective fusion of vision and dialog via attention visualization~(\cref{ssec:attn_vis}). 

\begin{table}[ht]
\centering
\resizebox{0.49\textwidth}{!}{
\begin{tabular}{p{0.03\textwidth}lrrrrrr}
\toprule
&\textbf{Model} & NDCG$\uparrow$ & MRR$\uparrow$ & R@1$\uparrow$ & R@5$\uparrow$ & R@10$\uparrow$ & Mean $\downarrow$\\
\midrule
\multirow{18}{*}{\parbox[t]{2mm}{\rotatebox[origin=c]{90}{Published Results}} $\begin{dcases*} \\ \\ \\ \\ \\ \\ \\ \\ \\ \\ \\ \\ \\ \\ \\  \end{dcases*}$}  
&NMN  &58.10 &58.80 &44.15 &76.88 &86.88 &4.81\\
&CorefNMN & 54.70 & 61.50 & 47.55 & 78.10 & 88.80 & 4.40\\
&GNN & 52.82 & 61.37 & 47.33 & 77.98 & 87.83 & 4.57\\ 
&FGA & 52.10 &63.70 &49.58 &80.97& 88.55 &4.51 \\
& DVAN & 54.70 &62.58 &48.90 &79.35 &89.03 &4.36\\
&RvA & 55.59 & 63.03 & 49.03 & 80.40 & 89.83 & 4.18\\ 
&DualVD & 56.32 &  63.23 & 49.25 & 80.23 & 89.70 & 4.11 \\
&HACAN & 57.17 &64.22 & 50.88 & 80.63 & 89.45 & 4.20\\ 
&Synergistic & 57.32 & 62.20 & 47.90 & 80.43 & 89.95 & 4.17\\
&Synergistic$^\dag$ &57.88 &63.42 &49.30 &80.77 &\underline{90.68} &3.97 \\
&DAN & 57.59 & 63.20 & 49.63 & 79.75 & 89.35 & 4.30\\ 
&DAN$^\dag$  &59.36 & \underline{64.92} &51.28 &\underline{81.60} &\textbf{90.88} &\underline{3.92}\\
&ReDAN$^\dag$  &  64.47 &53.73 &42.45 &64.68 &75.68 &6.64 \\
&CAG & 56.64 &63.49 &49.85 &80.63 &90.15 &4.11 \\
&Square$^\dag$  &60.16 &61.26 &47.15 &78.73& 88.48 &4.46 \\
&MCA$^*$ &72.47 &37.68 &20.67 & 56.67 &72.12& 8.89\\
&MReal-BDAI$^{\dag*}$ & 74.02 & 52.62 & 40.03 & 68.85 & 79.15 & 6.76 \\
&P1\_P2$^{\dag*}$  &\underline{74.91}&	49.13&36.68&62.98&78.55&7.03 \\
\midrule
\multirow{11}{*}{\parbox[t]{2mm}{\rotatebox[origin=c]{90}{Leaderboard Results}} $\begin{dcases*} \\ \\ \\ \\ \\ \\ \\ \\ \\ \end{dcases*}$}  
&LF & 45.31 & 55.42 & 40.95 & 72.45 & 82.83 & 5.95\\ 
&HRE & 45.46 & 54.16 & 39.93 & 70.45 & 81.50 & 6.41\\ 
&MN & 47.50 & 55.49 & 40.98 & 72.30 & 83.30 & 5.92\\ 
&MN-Att & 49.58 & 56.90 & 42.42 & 74.00 & 84.35 & 5.59\\ 
&LF-Att & 49.76 & 57.07 & 42.08 & 74.82 & 85.05 & 5.41\\ 
&MS ConvAI &55.35 &63.27 &49.53 &80.40 &89.60 &4.15\\
&UET-VNU$^\dag$ &57.40 &59.50 &45.50 &76.33 &85.82 &5.34 \\
&MVAN & 59.37 &64.84 &\underline{51.45} &81.12 &90.65 &3.97 \\
&SGLNs$^{\dag}$ & 61.27 &59.97 &45.68 &77.12 &87.10 &4.85 \\
&VisDial-BERT$^*$ & 74.47 & 50.74 & 37.95&64.13&80.00 &	6.28 \\
&Tohoku-CV$^{\dag*}$  &	74.88&52.14&38.93&66.60 &80.65&6.53 \\
\midrule

\multirow{3}{*}{\parbox[t]{2mm}{\rotatebox[origin=c]{90}{Ours}} $\begin{dcases*} \\ \\  \end{dcases*}$} 
&VD-BERT & 59.96 & \textbf{65.44} & \textbf{51.63} & \textbf{82.23} & \underline{90.68} & \textbf{3.90} \\
&VD-BERT$^*$ & 74.54& 46.72&33.15&61.58&77.15&7.18 \\
&VD-BERT$^{\dag*}$ &  \textbf{75.35} & 51.17 & 38.90  & 62.82 & 77.98 & 6.69\\
\bottomrule
\end{tabular}
}
\vspace{-0.5em}
\caption{Summary of results on the test-std split of VisDial v$1.0$ dataset. The results are reported by the test server.
``$\dag$'' denotes ensemble model and ``$*$'' indicates fine-tuning on dense annotations. The  ``$\uparrow$'' denotes higher value for better performance and ``$\downarrow$'' is the opposite. The best and second-best results in each column are in bold and underlined respectively.}\label{tabs:main_v10_full}
\vspace{-0.5em}
\end{table}

\subsection{Main Results}\label{ssec:main_res}
We report main quantitative comparison results on both VisDial v$1.0$ and v$0.9$ datasets below.
\vspace{-0.2em}
\paragraph{Comparison.} 
We  consider state-of-the-art published baselines, including   NMN~\cite{DBLP:conf/iccv/HuARDS17}, CorefNMN~\cite{DBLP:conf/eccv/KotturMPBR18},
GNN~\cite{DBLP:conf/cvpr/ZhengWQZ19},
FGA~\cite{DBLP:conf/cvpr/SchwartzYHS19},
DVAN~\cite{DBLP:conf/ijcai/GuoWW19},
RvA~\cite{DBLP:conf/cvpr/NiuZZZLW19}, 
DualVD~\cite{DBLP:conf/aaai/JiangYQZZ0W20},
HACAN~\cite{DBLP:conf/iccv/YangZZ19},
Synergistic~\cite{DBLP:conf/cvpr/GuoXT19},  DAN~\cite{DBLP:conf/emnlp/KangLZ19}, ReDAN~\cite{DBLP:conf/acl/GanCKLLG19},
CAG~\cite{DBLP:conf/cvpr/GuoWZZW20},
Square~\cite{DBLP:conf/aaai/KimTB20},
MCA~\cite{DBLP:conf/acl/AgarwalBLKR20}, MReal-BDAI and  P1\_P2~\cite{DBLP:conf/cvpr/QiNHZ20}.
We further report results from the
leaderboard\footnote{\url{https://evalai.cloudcv.org/web/challenges/challenge-page/161/leaderboard/483\#leaderboardrank-1}} for a more up-to-date comparison, where some   can be found in the arXiv, such as 
MVAN~\cite{DBLP:journals/corr/abs-2004-14025},
SGLNs~\cite{DBLP:journals/corr/abs-2004-06698},
VisDial-BERT~\cite{DBLP:journals/corr/abs-1912-02379}, and Tohoku-CV~\cite{DBLP:journals/corr/abs-1911-11390}.

\paragraph{Results on VisDial v1.0 test-std.}
We  report the  comparison results on VisDial v$1.0$ test-std split in Table~\ref{tabs:main_v10_full} and make the following observations.

\noindent$\bullet$~\textit{New state of the art for both single-model  and ensemble settings.}
Our single-model VD-BERT significantly outperforms all of its single-model counterparts across various metrics, even including some ensemble variants such as Synergistic, DAN (except R@10), and ReDAN (except NDCG).
With further fine-tuning on dense annotations, the NDCG score increases quite sharply, from $59.96$ to $74.54$ with nearly $15\%$ absolute improvement, setting a new state of the art in the single-model setting. 
This indicates that dense annotation fine-tuning plays a crucial role in boosting the NDCG scores. 
Moreover, 
our designed ensemble version yields new state of the art  ($\textbf{75.35}$ NDCG), outperforming the 2019 VisDial challenge winner MReal-BDAI ($74.02$ NDCG) by over $1.3$ absolute points. 

\begin{table}
\centering
\resizebox{0.49\textwidth}{!}{
\begin{tabular}{lccccc}
\toprule
\multirow{2}{*}{\textbf{Model}}  & MRR$\uparrow$ & R@1$\uparrow$ & R@5$\uparrow$ & R@10$\uparrow$ & Mean $\downarrow$ \\ 
\cmidrule(lr){2-6}
& \multicolumn{5}{c}{Discriminative/Generative} \\

\midrule
LF & 58.07/51.99& 43.82/41.83 & 74.68/61.78 & 84.07/67.59 & 5.78/17.07 \\
HRE & 58.46/52.37 &44.67/42.29 &74.50/62.18 &84.22/67.92 &5.72/17.07\\
HREA & 58.68/52.42& 44.82/42.28& 74.81/62.33 & 84.36/68.17& 5.66/16.79 \\
MN   & 59.65/52.59  & 45.55/42.29  & 76.22/62.85  & 85.37/68.88   & 5.46/17.06 \\
HCIAE & 62.22/54.67  & 48.48/44.35  & 78.75/65.28  & 87.59/71.55  & 4.81/14.23 \\
CoAtt & 63.98/55.78  & 50.29/46.10  & 80.71/\textbf{65.69}  & 88.81/71.74  & 4.47/14.43  \\
RvA   & 66.34/55.43  & 52.71/45.37  & \underline{82.97}/65.27  & \underline{90.73}/\textbf{72.97} & \textbf{3.93}/\textbf{10.71}\\
DVAN & \underline{66.67}/\underline{55.94} & \underline{53.62}/\underline{46.58} & 82.85/\underline{65.50} & 90.72/71.25 & \textbf{3.93}/14.79\\

\midrule
VD-BERT & \textbf{70.04}/\textbf{55.95} &\textbf{57.79}/\textbf{46.83} &\textbf{85.34}/65.43 &\textbf{92.68}/\underline{72.05} &\underline{4.04}/\underline{13.18} \\
\bottomrule
\end{tabular}
}
\vspace{-0.5em}
\caption{Discriminative and generative results of various models on the val split of VisDial v$0.9$ dataset. }\label{tabs:main_v09}
\vspace{-0.5em}
\end{table}

\noindent$\bullet$~\textit{Inconsistency between NDCG and other metrics.}
While dense annotation fine-tuning yields huge improvements on NDCG,
we also notice that it has a severe countereffect on other metrics, e.g., reducing the MRR score from $65.44$ to $46.72$ for VD-BERT.
Such a phenomenon has also been observed in other recent models, such as MReal-BDAI, VisDial-BERT, Tohoku-CV Lab, and P1\_P2, whose NDCG scores surpass others without dense annotation fine-tuning by at least around $10\%$ absolute points while other metrics drop dramatically.
We provide a detailed analysis of this phenomenon in \cref{ssec:dense}.

\noindent$\bullet$~\textit{Our VD-BERT is simpler and more effective than VisDial-BERT.}
VisDial-BERT is a concurrent work to ours that also exploits  vision-language pretrained models for visual dialog.
It only reports the single-model performance of $74.47$ NDCG. Compare to that, our VD-BERT achieves slightly better results ($74.54$ NDCG), however, note that we did not pretrain on large-scale external vision-language datasets like Conceptual Captions~\cite{DBLP:conf/acl/SoricutDSG18} and VQA as VisDial-BERT does. 
Besides, while VisDial-BERT does not observe improvements by ensembling, we endeavor to design an effective ensemble strategy (see Table~\ref{tabs:ablation}(d)) to increase the NDCG score to $75.35$ for VD-BERT.

\begin{table}
\centering
\resizebox{0.49\textwidth}{!}{
\begin{tabular}{llrrrrrr}
\toprule
&\textbf{Model} & NDCG$\uparrow$ & MRR$\uparrow$ & R@1$\uparrow$ & R@5$\uparrow$ & R@10$\uparrow$ & Mean $\downarrow$\\
\midrule
\multirow{4}{*}{(a)} 
&From scratch &56.20& 62.25 &48.16 &79.57&89.01& 4.31  \\
&Init from VLP &61.79&66.67&53.23&83.60&91.97&3.66  \\
&Init from BERT &\textbf{63.22} & \textbf{67.44} & \textbf{54.02} & \textbf{83.96} & \textbf{92.33} & \textbf{3.53}  \\
&~~$\hookrightarrow$ only NSP  &55.89 &63.15&48.98&80.45&89.72&4.15 \\

\midrule
\midrule

\multirow{4}{*}{(b)} 
&No history & \textbf{64.70} & 62.93 & 48.70 &  80.42 & 89.73 & 4.30 \\
&One previous turn & 63.47 & 65.30 & 51.66 & 82.30 & 90.97 &3.86 \\
&Full history & 63.22 & \textbf{67.44} & \textbf{54.02} & \textbf{83.96} & \textbf{92.33} & \textbf{3.53} \\
&~~$\hookrightarrow$ only text &54.32&62.79&48.48&80.12&89.33&4.27 \\

\midrule
\midrule
\multirow{4}{*}{(c)} 
&CE &74.47& 44.94 &32.23 &60.10&76.70& 7.57  \\
&ListNet &\textbf{74.54}&46.72&33.15&61.58&77.15&\textbf{7.18}  \\
&ListMLE &72.96 & 36.81 & 20.70 & 54.60 & 73.28 & 8.90  \\
&ApproxNDCG &72.45 &\textbf{49.88}&\textbf{37.88}&\textbf{62.90}&\textbf{77.40}&7.26 \\

\midrule
\midrule

\multirow{4}{*}{(d)} 
&\textsc{Epoch} & 74.84 & 47.40 & 34.30 & 61.58 & 77.78 & 7.12  \\
&\textsc{Length} & 75.07 & 47.33 & 33.88 & 62.20 & \textbf{78.50} & 7.01 \\
&\textsc{RANK} &  75.13 & 50.00 & 38.28  & 60.93 & 77.28 & 6.90\\
&\textsc{Diverse} &  \textbf{75.35} & \textbf{51.17} & \textbf{38.90}  & \textbf{62.82} & 77.98 & \textbf{6.69}\\
\bottomrule

\end{tabular}
}
\vspace{-0.5em}
\caption{Extensive ablation studies: training with (a) various  settings and (b)  contexts on v$1.0$ val; dense annotation fine-tuning with (c) varying ranking methods and (d) various ensemble strategies on v$1.0$ test-std.}\label{tabs:ablation}
\vspace{-0.5em} 
\end{table}

\paragraph{Results on VisDial v0.9 val.}
We further show both discriminative and generative results on v$0.9$ val split in Table~\ref{tabs:main_v09}. 
For comparison, we choose LF, HRE, HREA, MN~\cite{DBLP:conf/cvpr/DasKGSYMPB17}, HCIAE~\cite{DBLP:conf/nips/LuKYPB17}, CoAtt~\cite{DBLP:conf/cvpr/Wu0S0H18},  RvA, and DVAN as they contain results in both settings on the v$0.9$ val. 
These models employ dual decoders for each setting separately.
Our model continues to yield much better results  in the discriminative setting (e.g., $70.04$ MRR compared to DVAN's $66.67$) and  comparable results with the state of the art in the generative setting (e.g., $55.95$ MRR score vs. DVAN's $55.94$). This validates the effectiveness of our VD-BERT in both settings using a unified Transformer encoder. By contrast, VisDial-BERT can only support the discriminative setting.

\subsection{Ablation Study}\label{ssec:analysis}
We first study how different training settings influence the results in Table~\ref{tabs:ablation}(a).
We observe that initializing the model with weights from BERT indeed benefits the visual dialog task a lot, increasing the NDCG score by about $7\%$ absolute over the model trained from scratch. Surprisingly, the model initialized with the weights from VLP that was pretrained on Conceptual Captions~\cite{DBLP:conf/acl/SoricutDSG18}, does not work better than the one initialized from BERT. It might be due to the domain discrepancy between image captions  and  multi-turn dialogs, as well as the slightly different experiment settings
(e.g., we extract $36$ objects from image compared to their $100$ objects). 
Another possible reason might be that the VisDial data with more than one million image-dialog turn pairs 
can provide adequate contexts to adapt BERT for effective vision and dialog fusion.
We also find that the visually grounded MLM is crucial for transferring BERT into the multimodal setting, indicated by a large performance drop when using only NSP. 

We then examine the impact of varying the dialog context used for training in Table~\ref{tabs:ablation}(b).
With longer dialog history (``Full history''), our model indeed yields better results in most of the ranking metrics, while the one without using any dialog history obtains the highest NDCG score. This indicates that dense relevance scores might be annotated with less consideration of dialog history.
If we remove the visual cues from the ``Full history'' model, we see a drop in all metrics, especially, on NDCG. 
However, this version still  obtains comparable results to the ``No history'' variant, revealing that textual information dominates the VisDial task.

In Table~\ref{tabs:ablation}(c), we compare Cross Entropy (CE) training with a bunch of other listwise ranking optimization methods: ListNet~\cite{DBLP:conf/icml/CaoQLTL07}, ListMLE~\cite{DBLP:conf/icml/XiaLWZL08}, and approxNDCG~\cite{DBLP:journals/ir/QinLL10}. Among these methods, ListNet yields the best NDCG and Mean Rank, while the approxNDCG achieves the best MRR and Recall on VisDial v1.0 test-std. Therefore, we employ the ListNet as our ranking module.

We also explore ways to achieve the best ensemble performance with various model selection criteria in Table~\ref{tabs:ablation}(d).
We consider three criteria,  \textsc{Epoch}, \textsc{Length}, and \textsc{Rank} that respectively refer to predictions from different epochs of a single model, from different models trained with varying context lengths and with different ranking methods in Table~\ref{tabs:ablation}(b)-(c).
We use four predictions from each criterion and combine their diverse predictions (\textsc{diverse}) by summing up their normalized ranking scores.
We observe that \textsc{epoch} contributes the least to the ensemble performance while \textsc{rank} models are more helpful than \textsc{length} models.
The diverse set of them leads to the best performance.

\begin{figure}
\centering
\includegraphics[trim={0cm 1cm 0 0.2cm},scale=0.46]{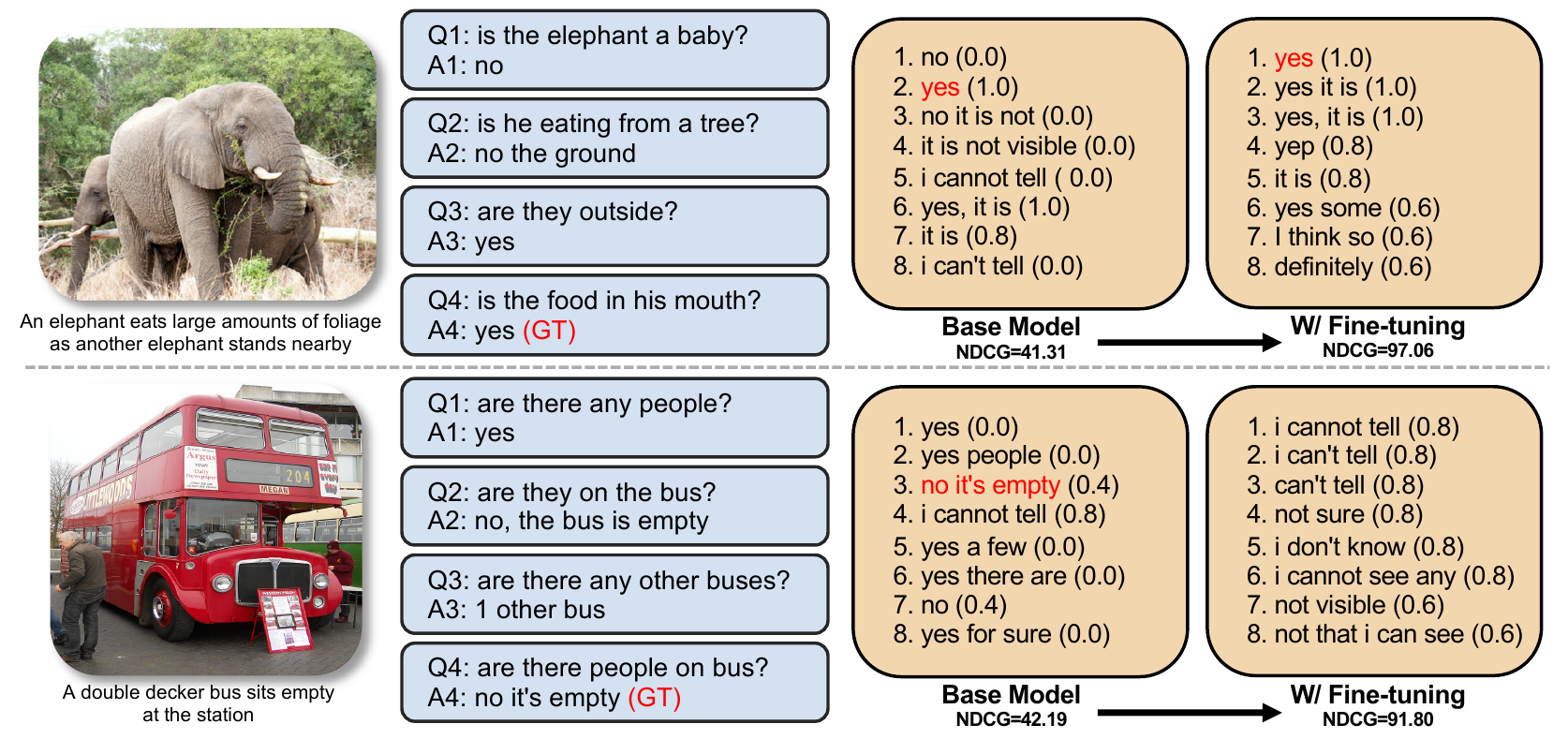}
\vspace{-0.5em}
\caption{The effects of dense annotation fine-tuning in our VD-BERT for two examples. GT: ground truth.}\label{fig:case}
\vspace{-1em}
\end{figure}

\begin{table*} [t]
\begin{center}
\resizebox{0.99\textwidth}{!}{

\begin{tabular}{cc}
\begin{minipage}{0.37\textwidth}
\centering
\captionsetup{type=figure}
\includegraphics[trim={0.5cm 0.5cm 0 0.5cm},scale=0.5]{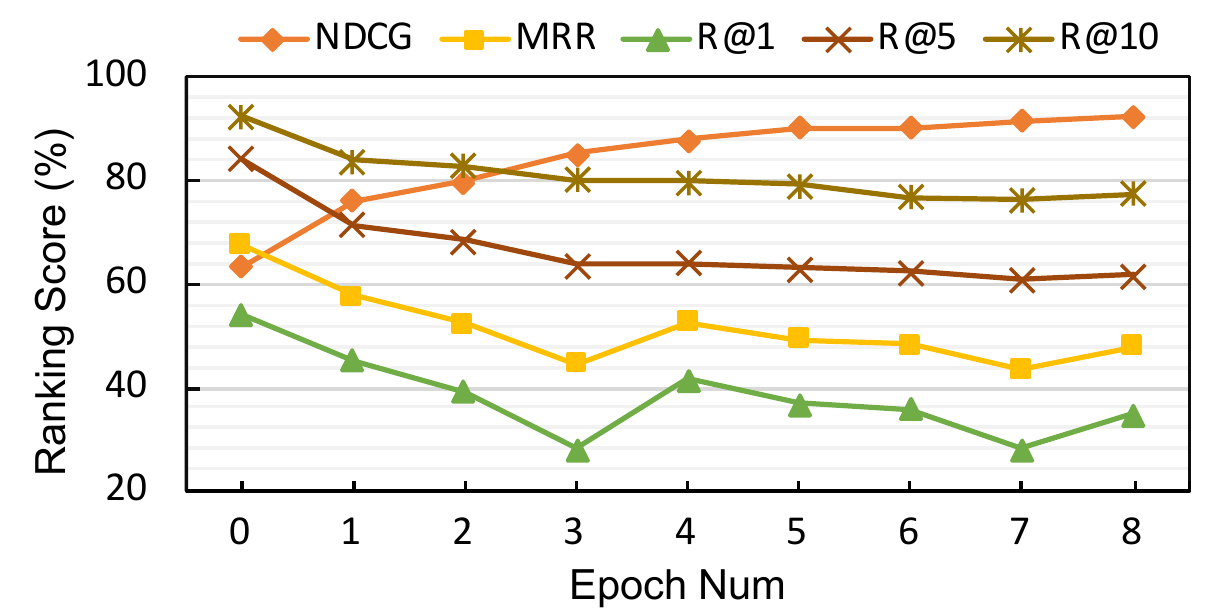}
\caption{Dense annotation fine-tuning on various metrics with the ListNet method.}\label{fig:rank_line}
\end{minipage}

&

\begin{minipage}{0.63\textwidth}
\resizebox{\textwidth}{!}{
\begin{tabular}{lccccccccc}
\toprule
\multirow{3}{*}{ \textbf{Models} } & \multirow{3}{*}{All} & \multicolumn{4}{c}{\textbf{Relevance Score}} &
\multicolumn{4}{c}{\textbf{Question Type}} \\
\cmidrule(lr){3-6}
\cmidrule(lr){7-10}
& & 1.0  & 0.6$\sim$0.8 & 0.2$\sim$0.4 & 0.0 & Yes/no  & Number & Color & Others   \\ 
&  & (31$\%$) & (35$\%$) & (25$\%$)& (9$\%$)  & (76$\%$) & (3$\%$) & (11$\%$) & (10$\%$) \\
\midrule
DAN & \multicolumn{1}{c|}{58.28} & 63.29&61.02 &53.29&  \multicolumn{1}{c|}{43.86} & 59.86 & 41.03 & 57.55 & 51.89  \\
Ours & \multicolumn{1}{c|}{63.55} & 70.25&65.18&58.40 & \multicolumn{1}{c|}{48.07}& 65.45 & 48.98 & 58.51 &58.75 \\
Ours (w/ ft) &  \multicolumn{1}{c|}{89.62} &95.38&89.76 &84.63& \multicolumn{1}{c|}{82.84}& 91.05 & 74.41  & 84.00 & 89.12 \\ 
\bottomrule
\end{tabular}
}
\caption{NDCG scores in VisDial v$1.0$ val split broken down into $4$ groups based on either the relevance score or the question type. The $\%$ value in the parentheses denotes the corresponding data proportion.}\label{tabs:rel_ques_type}
\end{minipage}
\end{tabular}
}
\end{center}
\vspace{-0.5em}
\end{table*}

\subsection{Fine-tuning on Dense Annotations}\label{ssec:dense}
In this section, we focus on the effect of dense annotation fine-tuning and try to analyze the reason of the inconsistency issue between NDCG and other ranking metrics (see Table~\ref{tabs:main_v10_full}) in the following.

\begin{figure*}[ht]
\centering
\includegraphics[width=0.99\textwidth]{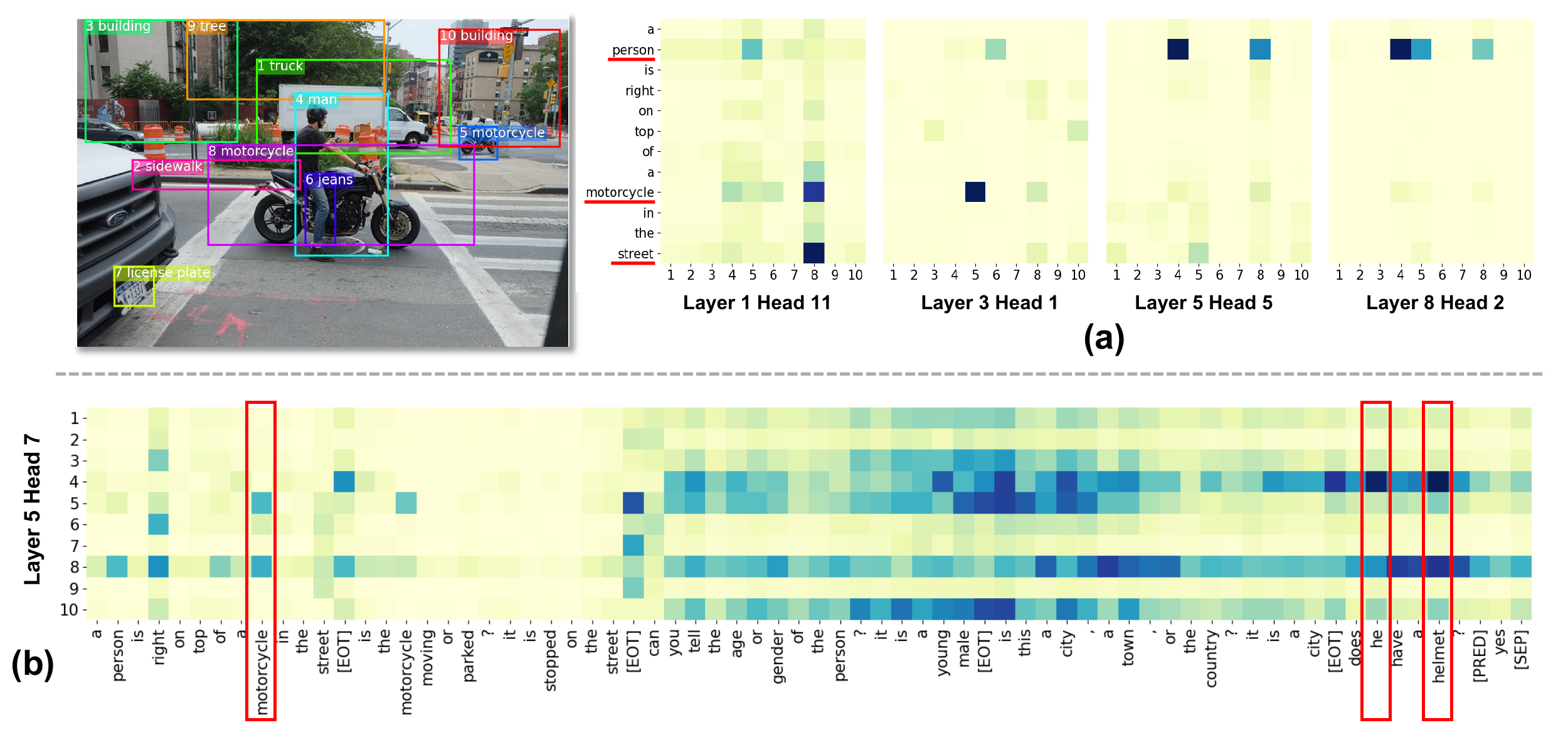}
 \vspace{-0.3em}
\caption{Attention weight visualization in our VD-BERT for a sampled image-dialog example.}\label{fig:attn_vis}
 \vspace{-0.3em}
\end{figure*}
\vspace{-0.2em}
\paragraph{Case Study.} 
We provide two examples to qualitatively demonstrate how dense annotation fine-tuning results in better NDCG scores in Figure~\ref{fig:case}. For the example at the top, fine-tuning helps our model to assign higher ranks to the answers that share similar semantics with the ground truth answer and should also be regarded as correct (``yes, it is'' and ``yep'' vs.  ``yes''). In the example at the bottom, we spot a mismatch between the sparse and dense annotations: the ground truth answer ``no, it's empty'' is only given a $0.4$ relevance score, while uncertain answers like ``i don't know'' are considered to be more relevant. 
In this case, fine-tuning instead makes our model fail to predict the correct answer despite the increase of  NDCG score.

\vspace{-0.2em}
\paragraph{Relevance Score and Question Type Analysis.}
We first show how various metrics change for fine-tuning in Figure~\ref{fig:rank_line}. For this experiment,  we randomly sample $200$ instances from VisDial v$1.0$ val as the test data and use the rest for fine-tuning with the ListNet ranking method. We observe that NDCG keeps increasing with more epochs of fine-tuning, while other  metrics such as Recall@K and MRR) drop. 
For further analysis, we classify the $2,064$ instances in VisDial v1.0 val set based on the ground-truth's relevance score and question type (Table~\ref{tabs:rel_ques_type}). We consider four bins $\{0.0, 0.2\sim0.4, 0.6\sim0.8, 1.0\}$ for the relevance score and four question types: \textit{Yes/no}, \textit{Number}, \textit{Color}, and \textit{Others}. We then analyze the NDCG scores assigned by DAN~\cite{DBLP:conf/emnlp/KangLZ19} and our VD-BERT with and without dense annotation fine-tuning. We choose DAN as it achieves good NDCG scores (Table~\ref{tabs:main_v10_full}) and provides the source code to reproduce their predictions.

By examining the distribution of the relevance scores, we find that only $31\%$ of them are aligned well with the sparse annotations and $9\%$ are totally misaligned.
As the degree of such mismatch increases (relevance score changes $1.0\rightarrow0.0$), both DAN and our model witness a plunge in NDCG ($63.29\rightarrow43.86$ and $70.25\rightarrow48.07$), while dense annotation fine-tuning significantly boosts NDCG scores for all groups, especially for the most misaligned one ($48.07\rightarrow82.84$ for our model). {These results validate that the misalignment of the sparse and dense annotations is the key reason for the inconsistency between NDCG and other metrics.}

For question types, we observe that \textit{Yes/no} is the major  type ($76\%$)  and also the easiest one, while \textit{Number} is the most challenging and least frequent one ($3\%$). 
Our model outperforms DAN by over $10\%$ in most of the question types except \textit{Color}.  Fine-tuning on dense annotations gives our model huge improvements across all the question types, especially for \textit{Others} with over $30\%$ absolute gain.

\subsection{Attention Visualization}\label{ssec:attn_vis}
To interpret our VD-BERT, we visualize the attention weights on the top 10 detected  objects  from its caption in Figure~\ref{fig:attn_vis}(a). We observe that many heads at different layers can correctly ground some entities like \texttt{person} and \texttt{motorcycle} in the image, and even reveal some high-level semantic correlations such as \texttt{person}$\leftrightarrow$\texttt{motorcycle} (at L8H2)
and \texttt{motorcycle}$\leftrightarrow$\texttt{street} (at L1H11). Besides, heads at higher layers tend to have a sharper focus on specific  objects like the man and the motorcycles in the image.

Next, we  examine how our VD-BERT captures the interactions between image and multi-turn dialog.
In contrast to other vision-language tasks, visual dialog has a more complex multi-turn structure, thereby posing a hurdle for effective fusion. As shown in Figure~\ref{fig:attn_vis}(b), VD-BERT can ground entities and discover some object relations, e.g., \texttt{helmet} is precisely related to the {man} and the {motorcycle} in the image (see the rightmost red box). More interestingly, it can even  resolve visual pronoun coreference of \texttt{he} in the question to the {man} in the image  (see the middle red box). We provide more qualitative examples in Figure~\ref{fig:more_attn_vis} and~\ref{fig:more_cases}.

\section{Conclusion}
We have presented VD-BERT, a unified vision-dialog Transformer model that exploits the pretrained BERT language models for visual dialog. VD-BERT is capable of modeling  all the interactions between an image and a multi-turn dialog  within a single-stream Transformer encoder and enables the effective fusion of  features from both modalities via simple visually grounded training. Besides, it can either rank or generate answers seamlessly. Without pretraining on external vision-language datasets, our model establishes new state-of-the-art performance in the discriminative setting and  shows promising results in the generative setting on the visual dialog benchmarks. 
\section*{Acknowledgements}
We thank  Chien-Sheng Wu, Jiashi Feng, Jiaxin Qi, and our anonymous reviewers for their insightful feedback on our paper.
This work was partially supported by the Research Grants Council of the
Hong Kong Special Administrative Region, China (No. CUHK 14210717, General Research Fund; No. CUHK 2300174, Collaborative Research Fund).

\bibliography{emnlp2020}
\bibliographystyle{acl_natbib}

\newpage
\appendix

\begin{figure*}
\centering
\includegraphics[trim={0 0 0 1em},width=0.85\textwidth]{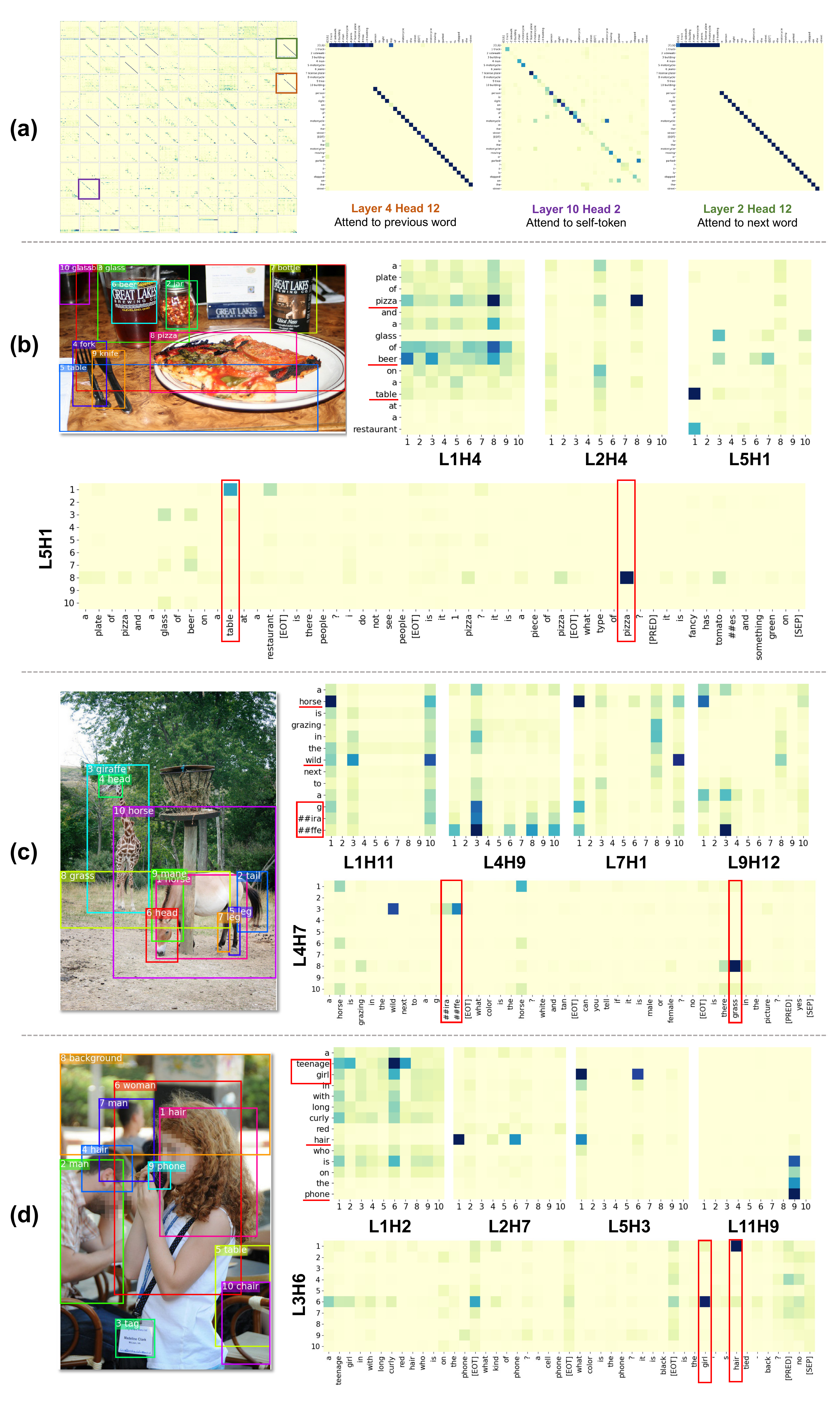}
\vspace{-0.5em}
\caption{More attention visualization examples showing that  VD-BERT achieves the effective fusion of vision and dialog contents. L$x$H$y$: Layer $x$ Head $y$ ($1\leq x,y \leq 12$). 
(a) It learns three apparent attention patterns for the example in Figure~\ref{fig:attn_vis}: attentions that a token puts to its previous token, to itself, and to the next token.
Besides, some of its attention heads can precisely ground some entities between image and caption/multi-turn dialog: (b) \texttt{pizza}, \texttt{beer}, and \texttt{table}; (c) \texttt{horse}, \texttt{wild}, and \texttt{giraffe}; (d) \texttt{teenage girl}, \texttt{hair}, and \texttt{phone}.}\label{fig:more_attn_vis}
\end{figure*}

\begin{figure*}
\centering
\includegraphics[scale=0.55]{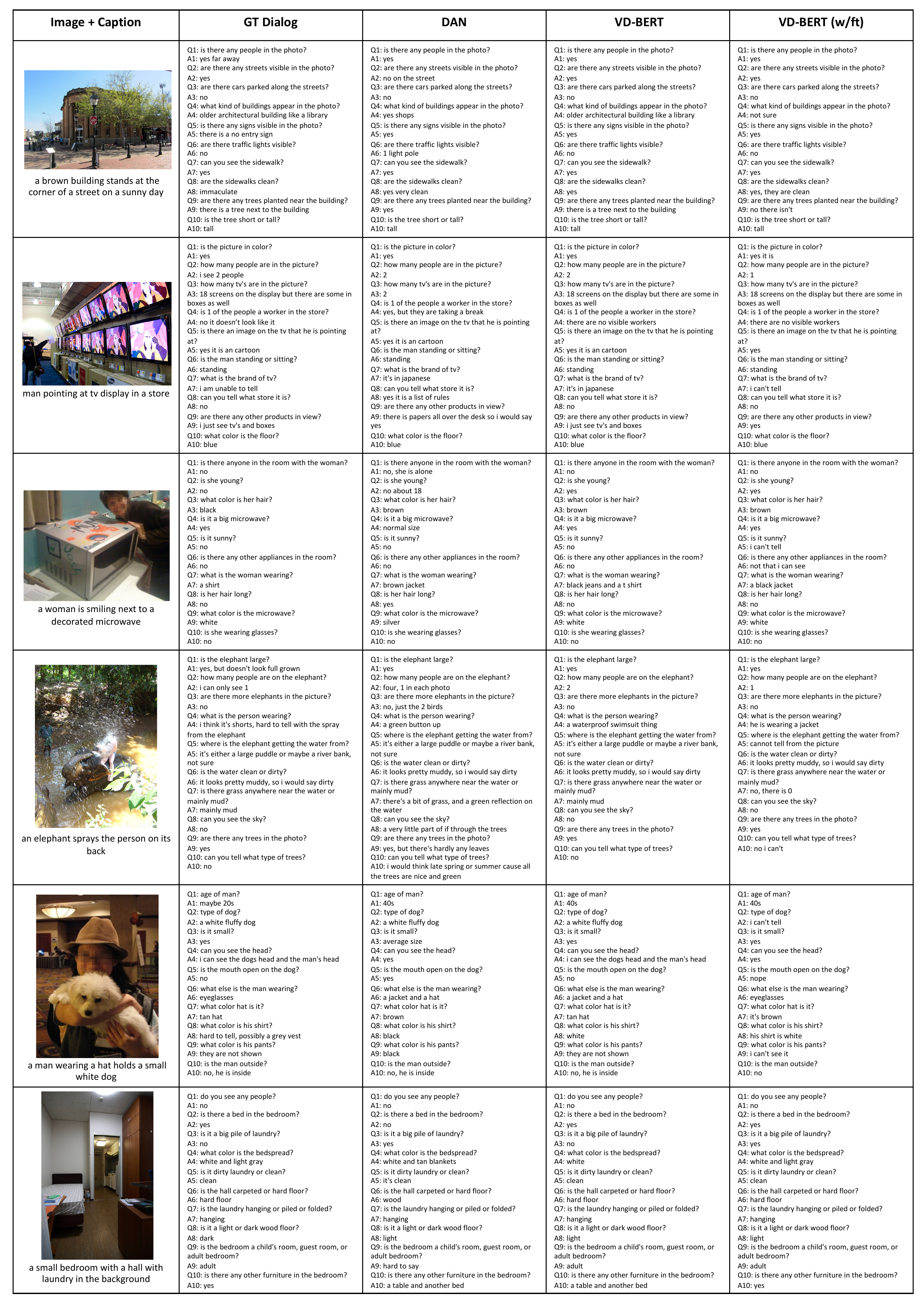}
\caption{More qualitative examples in VisDial v1.0 val split for three model variants: DAN~\cite{DBLP:conf/emnlp/KangLZ19}, VD-BERT, and VD-BERT with dense annotation fine-tuning. The second column is for ground truth (GT) dialog.}\label{fig:more_cases}
\end{figure*}

\end{document}